# Generalization algorithm of multimodal pre-training model based on graph-text self-supervised training

Xiaobing Zhang
College of Application and Technology
Shenzhen University
Shenzhen, China
149910819@qq.com

Zhenhao Tang
College of Application and Technology
Shenzhen University
Shenzhen, China
875378492@qq.com

Zi Long*
College of Big Data and Internet
Shenzhen Technology University
Shenzhen, China
longzi@sztu.edu.cn

Xianghua Fu
College of Big Data and Internet
Shenzhen Technology University
Shenzhen, China
fuxianghua@sztu.edu.cn

*Abstract*—Recently, a large number of studies have shown that the introduction of visual information can effectively improve the effect of neural machine translation (NMT). Its effectiveness largely depends on the availability of a large number of bilingual parallel sentence pairs and manual image annotation. The lack of images and the effectiveness of images have been difficult to solve. In this paper, a multimodal pre-training generalization algorithm for self-supervised training is proposed, which overcomes the lack of visual information and inaccuracy, and thus extends the applicability of images on NMT. Specifically, we will search for many pictures from the existing sentences through the search engine, and then through the relationship between visual information and text, do the self-supervised training task of graphics and text to obtain more effective visual information for text. We show that when the filtered information is used as multimodal machine translation for fine-tuning, the effect of translation in the global voice dataset is 0.5 BLEU higher than the baseline.

I. Introduction

Visual information plays a very important role in neural machine translation (NM)[1]. Generally speaking, visual information is related to sentence pairs. Through the end-to-end model framework, manually annotated visual information and bilingual text information are used to train multimodal NMT models, such as the public dataset Multi30k.

The advantage of multimodal neural machine translation[1] is that it can use visual information to improve the effect of machine translation. However, the effectiveness of visual information for translation depends to a large extent on its relevance to bilingual parallel sentence pairs, which hinders the applicability of images for machine translation. Some studies obtain relevant images in various ways, but the obtained image information may not be used in all data sets. Therefore, visual information is only applicable to the translation task of small and specific multi-model dataset Multi30K[2], but not to large-scale pure text NMT and low-resource pure text NMM, because the pictures corresponding to these texts are either not available, or the correlation with the text is too poor.

To solve these problems, Zhang et al. However, the subject image lookup table consists of a limited set of sentence image pairs, such as the Multi30K and MS COCO image caption data set[5]. Its image retrieval method is difficult to deal with words outside the vocabulary. Tang et al. proposed the method of searching keywords[4] to obtain relevant pictures to help the text obtain better translation effect. However, the obtained image information may not be used in all data sets. In addition, the language model of pre-training has been proved to be a useful tool for context representation. Many studies have found that cross-language and visual pre-training can be combined[3], and the introduction of these models can greatly speed up the training time and obtain better results.

In this article, we propose a multimodal pre-training generalization algorithm of self-supervised training, which uses the method of retrieving keywords to obtain multiple images and extract the visual information with strong relevance to the text from multiple images. It overcomes the shortcomings of previous visual information deficiency and inaccurate image information, thus expanding the applicability of images on NMT. At the same time, we use a cross-language and visual pre-training model to avoid the negative impact on the translation effect due to the lack of data in the text itself.

In summary, our contributions are as follows: (1) We propose to obtain more effective picture information through the task of self-supervised of pictures and texts, thus overcoming the shortcomings of insufficient or inaccurate picture information. (2). We have added a cross-language and visual pre-training model to avoid the negative impact on the translation effect due to the lack of data in the text itself. (3) The method we proposed is more universal for text or text errors and deficiencies, and thus more practical.[12]

## II. RELATED WORKS

In recent years, multimodal machine translation has gradually become the research hotspot and difficulty of machine translation, especially in the field of visual information. At the same time, the language model of pre-training has been proved to be a useful tool for context representation, and the combination of cross-language and visual pre-training is also quite innovative.

In view of the lack of image information in multimodal machine translation, Tang et al. proposed to obtain image information through search keyword algorithm to achieve better results. The focus layer with gating weighting is a UVR model that integrates visual information and text information as input into the decoder to predict target translation. For visual information and pre-training, a large-scale image caption corpus is used to induce basic visual and language expression, This is usually achieved by extending the masking language modeling (MLM) target [3]and assisting visual and language tasks (such as masking area classification and image sentence matching).

The VLTM model [3]demonstrates the effectiveness of the combination of cross-language and visual pre-training.

In summary, the above work still lacks effective visual information, and the use of cross-language and visual pre-training has never been used in the absence of visual information. Therefore, we propose a generalization algorithm of multimodal pre-training model based on graph-text self-supervised training.[17]

## III. MODEL AND METHODS

In this section, we will introduce the multimodal pre-training generalization algorithm model of self-supervised training, which is mainly composed of three parts: (1) text and image feature processing.[13] (2) Self-supervised task part of image and text. (3) Input the extracted more accurate image features into the model for machine translation fine-tuning.

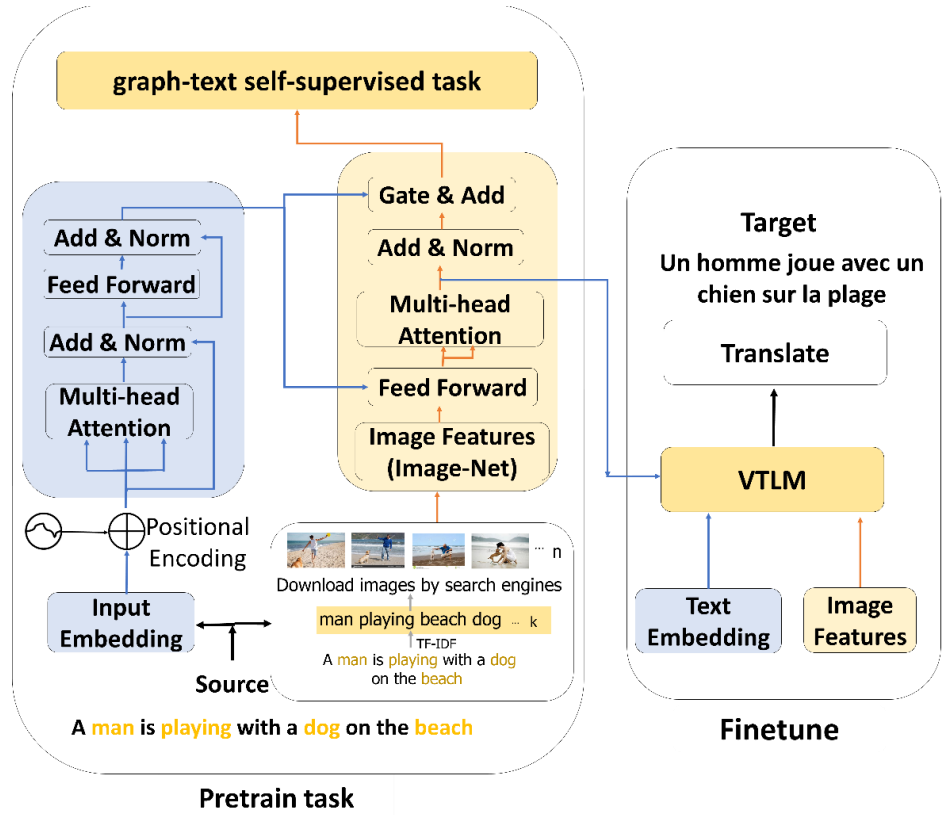

Fig. 1. Architectures Model

### A. Text and Image Embedding

After the text and pictures are obtained through the search engine, the next step is the coding problem. We use the similar Transfomer model to obtain the corresponding text information.

### B. Self-supervised Task

The image and text self-supervised task we use is to search for the corresponding relationship between the words and pictures. When the search words are in the original sentence, we mark them as 1, otherwise, we mark them as 0. In this way, we calculate the loss value of all values.[19]

About two weeks ago , Gonu , a tropical storm , hit Oman and the Sistan and Balouchestan province in Iran .
Keyword:About two weeks ago  Gonu
1 1 1 1 0 1 0 0 0 0 0 0 0 0 0 0 0 0 0 0 0 0 0 0

Fig. 2.  Seqence label

The above figure shows that the key words are extracted from the original sentence, the key words in the sentence are marked with 1 in the original sentence, and the key words that are not keywords are marked with 0. Finally, the cross entropy loss is calculated to get accuracy.

### C. Machine translation fine-tuning.

We input the image feature vector obtained from the self-supervised task into the fine-tuning model, and then use the pre-trained text image model to calculate the translation effect[9]. The fine-tuning model here is the VLTM model, which is the first model to use both text and image in the pre-training data. The model diagram is proposed in the VLTM paper.

## IV. EXPERIMENT AND RESULT ANALYSIS

### A. Data

Usually, the original visual and linguistic sentence is Multi30k, which is translated from English to German. Because we want to discuss the situation when visual information is insufficient, we use the data set from Global Voice English to German. The following is a detailed description of Global Voice.

Global Voices Global Voices (EN-DE) dataset consists of more than 70k bilingual sentence pairs from summaries of news articles. We randomly sample 2000 data as dev set, 2000 as test set, and use the remained as training set. As shown in Table 1, it is the amount of data corresponding to Global Voice[21] in training, verification and testing.

TABLE I.    DATASET DISTRIBUTION

| Languages | Training | Validation | Testing | Total |
|---|---|---|---|---|
| English | 67321 | 2104 | 2104 | 71529 |
| Germer | 67321 | 2104 | 2104 | 71529 |

### B. System Settings

The experimental steps of the system can be divided into image acquisition, text and image feature extraction, self-supervised  task of graph-text correlation, and fine-tuning VLTM model output.

Among them, the most important task is the image and text self-supervised task. In the image and text self-supervised task, we compress 36 regions of five images into 36 regions of one image, and remove the image farthest from the text.

### C. Experimental Results

For the task of sequence label image and text self-supervised , we have obtained 83% of the correct rate, which is about 33% higher than 50% of the random image and text irrelevant task. This shows that the image and text self-supervised  task is helpful for extracting effective visual information. We ran nearly 1000 epochs, and the accuracy rate began to rise with the increase of the number of epochs. After reaching 500, it maintained a basically unchanged trend, as shown in Figure 3.

| Epoch | 0 | 100 | 200 | 300 | 400 | 500 | 600 | 700 | 800 | 900 | 1000 |
|---|---|---|---|---|---|---|---|---|---|---|---|
| 系列 1 | 0.5 | 0.7 | 0.7 | 0.7 | 0.8 | 0.8 | 0.8 | 0.8 | 0.8 | 0.8 | 0.8 |

Fig. 3.  Accuray with different epoch

Finally, through self-supervised training related to graphics and texts, we extracted 36 key feature combinations that can more accurately describe the graphics of each bilingual pair. We input the vector combinations of the images into the VTLM model to obtain more effective The information is translated on the original text of the Global Voice dataset, and the BLEU value of the translation effect obtained is shown in the table below.

TABLE II.    EFFECT WITH DIFFERENT METHODS

| Global Voice | NMT | MMT |
|---|---|---|
| Tang et al(2022) | 9.22 | 9.81 |
| TLM | 21.80 | 21.70 |
| VTLM+ Image | 21.90 | 21.90 |
| TLM+ Image +SSP | / | 21.95 |
| VTLM+ Image +SSP | / | **22.40** |

Among them, when the VTLM pre-training model is not used, because the training text data is too small, the initial BLEU data is around 9[4];The data of TLM and TLM+Image is obtained by using the VTLM pre-training model and searching for pictures, which is 12 more BLEU values than the data without the pre-training model; SSP is the abbreviation of the image-text self-supervised model. After the self-supervision of the text, there is 0.5 more BLEU values than the original unprocessed pictures, which is the contribution of this article[11][13][18].
3

*D. Intuitive Explanation Example*

In order to intuitively understand why the selected image features improve the translation effect of the text. As shown in the comparison in the figure below, the original "raiding" should be translated into "griffen" instead of "hielt", which is consistent with the characteristics of the fourth picture.[19]

| | |
|---|---|
| Test(EN): | the police were **raiding** the building . |
| Test(DE): | die polizei **sturmte** unser gebaude . |
| TLM: | die polizei **hielt** das gebaude an. |
| VTLM+ Image +SSP: | die polizisten **griffen** das gebaude ein . |

Fig. 4. Example of sentence

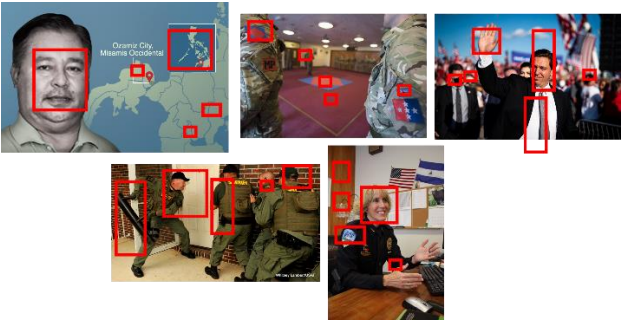

Fig. 5. Example of picture with the tagging1

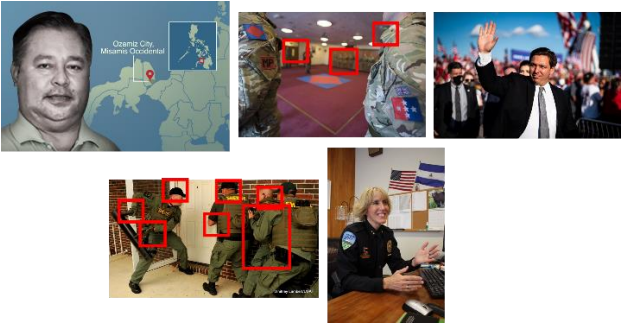

Fig. 6. Example of picture with the tagging2

All in all, the features that do not use graphic-text supervised learning are evenly distributed in the 5 pictures, while the features selected by graphic-text supervised learning focus on the fourth picture, which excludes the influence of other irrelevant information on translation. So as to achieve a good translation effect.

## V. CONCLUSION

In short, the image-text self-supervised training algorithm we proposed can concentrate more text-related visual [9][10][11]information and reduce irrelevant visual information, so as to achieve better translation results in machine translation; at the same time, using the VLTM pre-training model can also Make the text fully learned, thereby reducing the impact of the text itself on multimodal machine translation

In the future, we can apply this graphic-text self-supervised algorithm to unexpected tasks of machine translation, thus becoming a general-purpose algorithm.

ACKNOWLEDGMENT

I am very grateful to Mr. Long Zi for his help in this research experiment and thesis writing; At the same time, I also want to thank Shenzhen University for giving me a good learning environment; Finally, I would like to thank the expert judges of the 5th International Conference on Natural Language Processing Organizing Committee.